\documentclass{bmvc2k}

\usepackage{graphicx}
\usepackage{subfigure}
\usepackage{amsmath,amssymb} 
\usepackage{multirow}
\usepackage{caption}
\title{Tiny-DSOD: Lightweight Object Detection for Resource-Restricted Usages}

\addauthor{Yuxi Li}{lyxok1@sjtu.edu.cn}{1}
\addauthor{Jiuwei Li}{jiuwei.li@intel.com}{2}
\addauthor{Weiyao Lin}{wylin@sjtu.edu.cn}{1}
\addauthor{Jianguo Li}{jianguo.li@intel.com}{2}

\addinstitution{
 Shanghai Jiao Tong University, China
}
\addinstitution{
 Intel Lab China
}

\runninghead{Li \etal}{Tiny-DSOD}

\def\etal{\emph{et al}\bmvaOneDot}

\begin{document}

\maketitle

\begin{abstract}
 Object detection has made great progress in the past few years along with the development of deep learning.
However, most current object detection methods are resource hungry, which hinders their wide deployment to many resource restricted usages such as usages on always-on devices, battery-powered low-end devices, etc.
This paper considers the resource and accuracy trade-off for resource-restricted usages during designing the whole object detection framework.
Based on the deeply supervised object detection (DSOD) framework, we propose Tiny-DSOD dedicating to resource-restricted usages.
Tiny-DSOD introduces two innovative and ultra-efficient architecture blocks: depthwise dense block (DDB) based backbone and depthwise feature-pyramid-network (D-FPN) based front-end.
We conduct extensive experiments on three famous benchmarks (PASCAL VOC 2007, KITTI, and COCO), and compare Tiny-DSOD to the state-of-the-art ultra-efficient object detection solutions such as Tiny-YOLO, MobileNet-SSD (v1 \& v2), SqueezeDet, Pelee, etc.
Results show that Tiny-DSOD outperforms these solutions in all the three metrics (parameter-size, FLOPs, accuracy) in each comparison.
For instance, Tiny-DSOD achieves $72.1\%$ mAP with only 0.95M parameters and 1.06B FLOPs, which is by far
the state-of-the-arts result with such a low resource requirement.\footnote{Jianguo Li and Weiyao Lin are the co-correspondence authors}
\end{abstract}

\section{Introduction}
Object detection is considered as a crucial and challenging task in the field of computer vision, since it involves the combination of object classification
and localization within a scene. Accompanied with the development of modern deep learning technique, lots of convolutional neural network (CNN) based detection frameworks, including Faster R-CNN~\cite{Ren2015Faster},YOLO~\cite{Redmon2016You}, SSD~\cite{Liu2016SSD} and their variants \cite{Dai2016R,He2017Mask,Fu2017DSSD,Xiang2017Context,Lin2017Feature,Redmon2016YOLO9000}, have been proposed and greatly promote the accuracy
for object detection.

In spite of the state-of-art accuracy these models achieved, most of them are resource hungry as they have both high computing complexity and large parameter size (or large model size).
High computing complexity requires computing units with higher peak FLOPs, which usually increases the budget for power consumptions.
The speed and accuracy trade-off has been extensively studied in \cite{huang2016speed}.
However, resources are not only computing resources, but also memory resources.
Large model size yields large persistent memory requirement, which is not only costly, but also power inefficient for low-end embedding applications due to frequently persistent memory access.
Due to these two limitations, majority of current object detection solutions are not suitable for low-power using scenarios such as applications on always-on devices or battery-powered low-end devices.

To alleviate such limitations, recently many researches were dedicated to ultra-efficient object detection network design.
For instance, YOLO \cite{Redmon2016You} provides a lite version named Tiny-YOLO, which compress the parameter size of YOLO to $15$M and achieves a detection speed of more than 200 fps on PASCAL VOC 2007 dataset \cite{Everingham2010Pascal}.
SqueezeDet \cite{Wu2017SqueezeDet} introduces SqueezeNet \cite{Forrest2016Squeeze} based backbone into the YOLO framework for efficient autonomous driving usages.
MobileNet-SSD adopts MobileNet \cite{Howard2017MobileNets} as backbone in the SSD framework, which yield a model with only 5.5M parameters and 1.14B FLOPs of computing on PASCAL VOC 2007 dataset.
Although these small networks reduce the computation resource requirement to a large extent, there is still a large accuracy gap between small networks and the full-sized counterparts.
For instance, there is 9.2\% accuracy drop from SSD (77.2\%) to MobileNet-SSD (68.0\%) on PASCAL VOC 2007.
In a nutshell, these small detection networks are far from achieving a good trade-off between resources (FLOPs \& memory) and accuracy.

We propose Tiny-DSOD, dedicating for obtaining a good balance between the resources (FLOPs \& memory) and accuracy trade-off.
The backbone part of the proposed framework is inspired by the object detection work DSOD \cite{Shen2017DSOD} and recent ultra-efficient depthwise separable convolution network structures from \cite{Chollet2016Xception,Howard2017MobileNets, Sandler2018Inverted}.
DSOD \cite{Shen2017DSOD} introduces several important principles to train object detection network from scratch,
in which deep supervision is most critical to help back-propagating supervised information from the loss layer to shallower layers without the gradient vanishing problem.
Implicit deep supervision like DenseNet structure~\cite{Huang2017Densely} was adopted in DSOD.
Tiny-DSOD combines the ultra-efficient depthwise separable convolutions from \cite{Chollet2016Xception,Howard2017MobileNets, Sandler2018Inverted} into the DenseNet, and
introduces a novel depthwise dense block (DDB) to replace dense blocks in DenseNet.
This design will not only reduce the computing resource requirements, but also keep the implicit deep supervision for efficient training.
For the front-end part, we try to bring the successful feature-pyramid-network (FPN) \cite{Lin2017Feature} into our framework
for fusing semantic information from low-resolution scale to neighborhood high resolution scale seamlessly.
We invent depthwise FPN (D-FPN) by incorporating the efficient depth-wise convolution into FPN.
We verify in our experiment that the lightweight D-FPN front-end can boost the detection accuracy notably.

We conduct extensive experiments to verify the effectiveness of Tiny-DSOD on different datasets like PASCAL VOC \cite{Everingham2010Pascal}, KITTI \cite{Geiger2012KITTI} and COCO \cite{Lin2014Microsoft}.
The results show that our Tiny-DSOD achieves a much better trade-off between resources (FLOPs \& memory) and accuracy.
For instance, On PASCAL VOC2007, Tiny-DSOD achieves a mean average precision (mAP) of $72.1\%$ with only $0.95$M parameters and $1.06$B FLOPs of computing.
To the best of our knowledge, this is the first detection model that can achieve $>70\%$ mAP with parameter size less than $1.0$M.
Actually, Tiny-DSOD outperforms state-of-the-art ultra-efficient object detectors such as Tiny-YOLO, SqueezeDet, MobileNet-SSD
in the three compared benchmark datasets (VOC 2007, KITTI, COCO) on all the three evaluation metrics (accuracy, parameter-size, FLOPs).
Compared with the smallest DSOD model \cite{Shen2017DSOD}, Tiny-DSOD reduces the parameter size to about 1/6 and the computing FLOPs to 1/5 with only 1.5\% accuracy drops.
%
%
The contributions of this paper are summarized as follows:
\begin{itemize}
	\setlength{\topsep}{1pt}
	\setlength{\itemsep}{1pt}
	\setlength{\parskip}{1pt}
	\item We propose depthwise dense block (DDB), a novel and efficient network structure to combine depthwise separable convolution with densely connected networks (DenseNet) for ultra-efficient computer vision usages.
	\item We propose D-FPN, a novel and lightweight version of FPN \cite{Lin2017Feature}, to fuse semantic information from neighborhood scales for boosting object detection accuracy.
	\item We design the ultra-efficient object detector Tiny-DSOD for resource-restricted usages based on the proposed DDB and D-FPN blocks.
	Tiny-DSOD outperforms state-of-the-art ultra-efficient object detectors such as Tiny-YOLO, SqueezeDet, MobileNet-SSD, etc
	in each of the three compared benchmark datasets (VOC 2007, KITTI, COCO) on all the three evaluation metrics (accuracy, parameter-size, FLOPs).
\end{itemize}

\section{Related Works}
\noindent\textbf{\large{State-of-the-arts Object Detection Networks}}

\vspace{1ex}
A various of CNN based object detection frameworks have been proposed in the past few years along with the fast development of deep learning.
They could be generally divided into two categories: single-stage based methods and two-stage-based methods.

Typical two-stage methods include R-CNN~\cite{girshick2014rich}, Fast R-CNN~\cite{girshick2015fast}, Faster RCNN~\cite{Ren2015Faster} and R-FCN~\cite{Dai2016R}.
Early methods like R-CNN~\cite{girshick2014rich} and Fast R-CNN~\cite{girshick2015fast} utilize external region proposal generation algorithms like \cite{uijlings2013selective} to produce region proposal candidates and perform classification on each candidate region.
Latter methods introduce region proposal networks (RPN) to produce region proposal, and integrate backbone network, RPN and front-end modules like classification and bounding-box regression module into one framework for end-to-end training.
This kind of methods are accurate but with heavy computing cost, and thus yield slow processing speed.

On the contrary, typical single-stage methods like SSD~\cite{Liu2016SSD} and YOLO~\cite{Redmon2016You}, apply pre-defined sliding default boxes of different scales/sizes on one or multiple feature maps to achieve the trade-off between speed and accuracy. This kind of methods are usually faster than the two-stage counterparts, but less accurate than two-stage-based methods.

Moreover, all these mentioned detection frameworks achieve better detection accuracy with a better backbone network (like ResNet~\cite{He2015Deep} or VGG-16~\cite{Simonyan2014Very}) as feature extractor, which is over-parameterized and consumes much computation resource.

\vspace{1ex}
\noindent\textbf{\large{Lightweight Object Detection Networks}}

\vspace{1ex}
The speed and accuracy trade-off has been extensively studied in \cite{huang2016speed}.
However, resources are not only computing cost for speed, but also memory resources.
Recently, many efforts are dedicated to design efficient and small-size networks for object detection for resource-restricted usages.
SqueezeNet \cite{Forrest2016Squeeze} (a simple version of inception \cite{szegedy2015going} structure named Fire module) based backbone is recently introduced into the modern single-stage frameworks for efficient detection \cite{Wu2017SqueezeDet}, which achieves comparable results on PASCAL VOC 2007 and KITTI~\cite{Geiger2012KITTI}.
For instance, SqueezeNet based SSD achieves 64.3\% mAP with only 5.5M parameters and 1.18B FLOPs of computing on PASCAL VOC 2007.

Meanwhile, depthwise separable convolution \cite{Howard2017MobileNets,Sandler2018Inverted,Chollet2016Xception} has shown great parameter and computing efficiency in generic image classification tasks. It was also introduced into the SSD framework for object detection purpose as a backbone and named as MobileNet-SSD \cite{Howard2017MobileNets}.
MobileNet-SSD achieves 68.0\% mAP with only 5.5M parameters and 1.14B FLOPs of computing on PASCAL VOC 2007.

Pelee \cite{Wang2018Pelee} utilizes a two-way densely connected structure to reduce computation consumption while keeping detection accuracy for mobile applications.

Nevertheless, there is still a large accuracy gap between efficient yet tiny networks and those of full-sized counterparts.
For instance, tiny YOLO can achieve 57.1\% mAP on PASCAL VOC 2007, while YOLOv2~\cite{Redmon2016YOLO9000} can reach $78.6\%$ mAP under the same setting.
SqueezeNet-SSD and MobileNet-SSD achieve 64.3\% and 68.0\% mAP on PASCAL VOC 2007 respectively, while full SSD reaches 77.2\% mAP under the same setting.
This observation inspires us that there is still a large space to achieve a better trade-off between resources (FLOPs \& memory) and accuracy for designing object detection networks.

\section{Method}
Our goal is to design an ultra-efficient object detection networks towards resource-restricted usages.
Our detector is based on the single-shot detector (SSD) \cite{Liu2016SSD} framework and the deeply-supervised object detection (DSOD) framework \cite{Shen2017DSOD}, which consists of the backbone part and the front-end part. We will elaborate these two parts below separately.

\subsection{{Depthwise Dense Blocks Based Backbone}}\label{sec:backbone}

Inspired by DSOD~\cite{Shen2017DSOD},
we also construct a DenseNet-like \cite{Huang2017Densely} backbone since it is easier to be trained from scratch with relatively fewer training set.
Taking the restricted resource into consideration, we introduce the ultra-efficient depth-wise separable convolution into the typical dense block, and refer to this new structure unit as depth-wise dense block (DDB).

\begin{figure}[]
	\vspace{2ex}
	\centering
	\subfigure[stacked DDB-a(w, g)]{\label{fig:DDB-a}\includegraphics[width=0.485\textwidth]{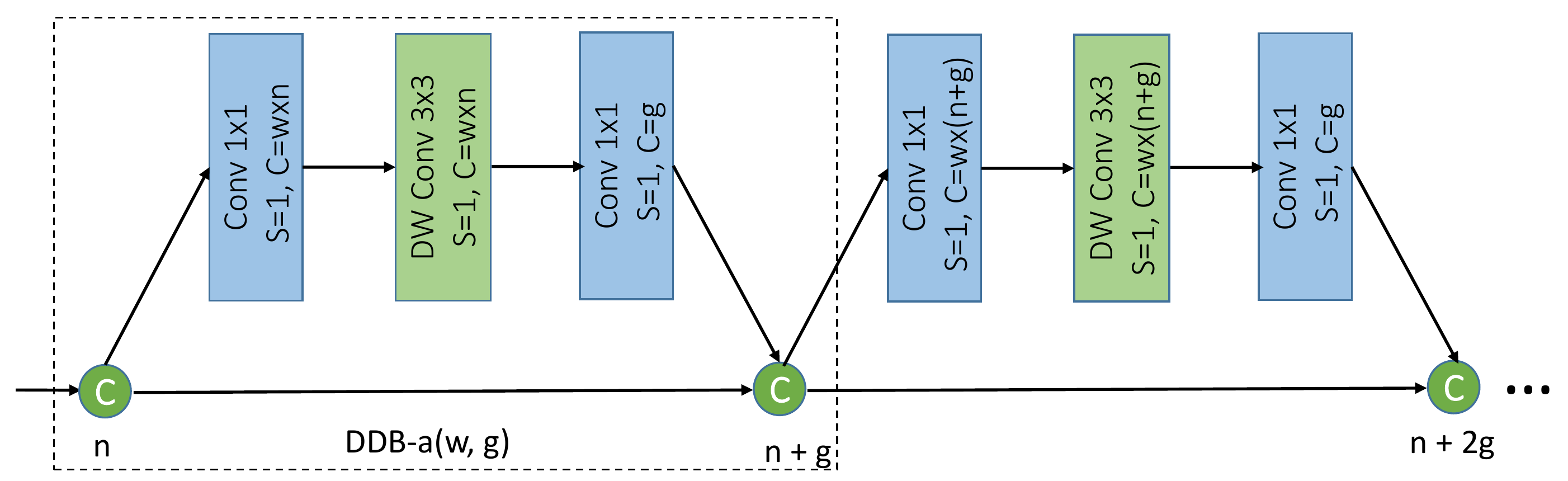}}
	\hspace{1ex}
	\subfigure[stacked DDB-b(g)]{\label{fig:DDB-b}\includegraphics[width=0.485\textwidth]{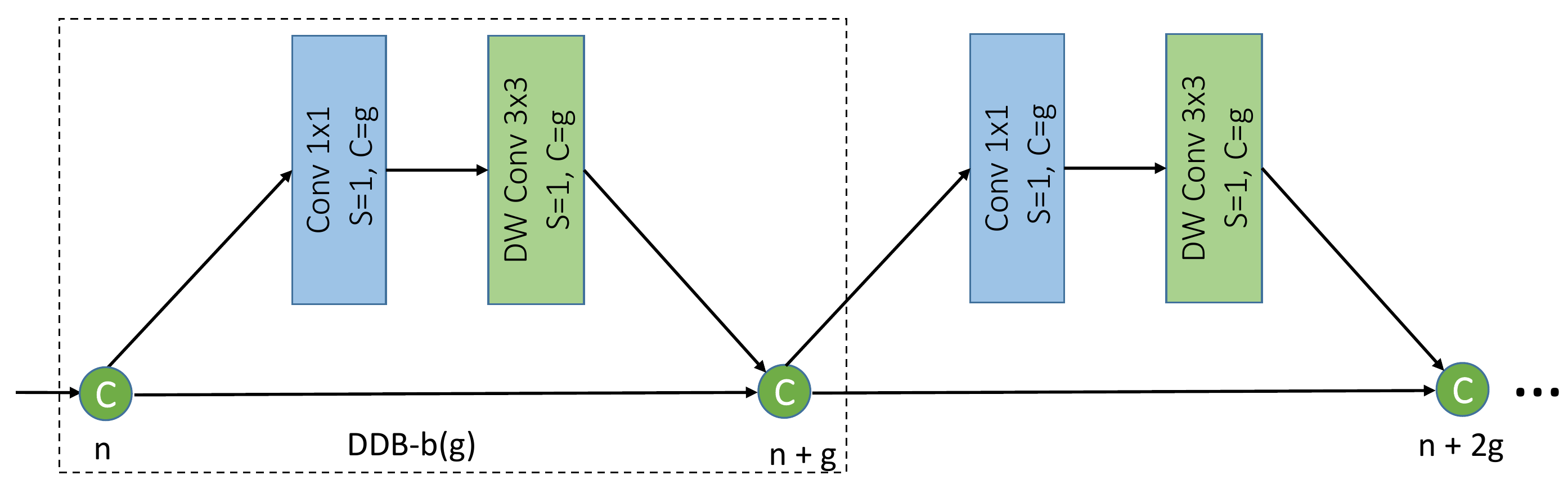}}
	\vspace{0.5ex}
	\caption{Illustrations of the depth-wise dense blocks (DDB). Two types of DDB are shown in the figure.
		In the rectangle, "S" means the stride of convolution, and "C" means the number of output channels.
		Numbers under the concatenating node (green C with circle) means the number of output channels after concatenation.
		(a) is stacked DDB-a parameterized by growth rate $g$ and expand ratio $w$. (b) is stacked DDB-b parameterized by growth rate $g$.}\label{fig:DDBs}
	\vspace{-3ex}
\end{figure}

We propose two types of DDB units, DDB-a and DDB-b, as shown in \autoref{fig:DDBs}.
The DDB-a unit in Figure \ref{fig:DDB-a} is inspired by the novel inverted residual blocks proposed in MobileNet-v2 \cite{Sandler2018Inverted}.
It first expands the input channels $w$ times to $w\times n$, where $n$ is the block input channel number, and $w$ is an integer hyper-parameter to control the model capacity.
It then applies the depth-wise convolution, and further projects feature maps to $g$ channels ($g$ is the growth rate of DDB-a) with a point-wise convolution (i.e., 1$\times$1 convolution). Finally, we use concatenation to merge the input and output feature maps together, instead of the residual addition operation in MobileNet-v2 \cite{Sandler2018Inverted}. DDB-a has two hyper-parameters $w$ and $g$, so we denote it as DDB-a$(w,g)$.

There are two main defects of DDB-a. \textit{First}, suppose $L$ DDB-a blocks are stacked, the complexity of the stacked structure is $O(L^3g^2)$.
This means the resource consumption grows rapidly with respect to $L$, so that
we have to control growth-rate $g$ to a small value even if just stacking several DDB-a together.
However, small growth rate $g$ will hurt the discriminate power of the whole model.
\textit{Second}, DDB-a concatenates the condensed (aka 1$\times$1 convolution projected) feature maps, so that there are continuous 1$\times$1 convolutions within two adjacent DDB-a units. This kind of processing will introduce potential redundancy among model parameters.

With this consideration, we design the other type of depth-wise dense block named DDB-b as shown in Figure \ref{fig:DDB-b}.
DDB-b first compresses the input channel to the size of growth rate $g$, and then perform depth-wise convolution.
The output of depthwise convolution is directly concatenate to the input without extra 1$\times$ 1 projection.
The overall complexity of $L$ stacked DDB-b blocks is $O(L^2g^2)$, which is smaller than that of DDB-a.
We will further verify by experiments in section \ref{sec:ablation} that DDB-b is not only more efficient but also more accurate than DDB-a under similar resource constraint.
Therefore we choose DDB-b as the basic unit to construct our final structure of backbone subnetwork.

\autoref{tab:backbone} shows the detailed structure of our backbone network. Each convolution layer is followed by a batch normalization and a ReLU layer.
There are four DDB stages in the Extractor part, where each DDB stage contains several DDB blocks,
followed by one transition layer to fuse channel-wise information from the last stage and compress the channel number for computing and parameter efficiency.
We also adopt the variational growth rate strategy in \cite{Huang2017CondenseNet}, by assigning a smaller $g$ to shallower stages with large spatial size, and increasing $g$ linearly when the stage goes deeper.
This will help saving computing cost since large spatial size at shallower stages usually consumes more computation.

\begin{table}[]
	\centering
	\footnotesize
	\begin{tabular}{c|c|c|c}\hline
		\multicolumn{2}{c|}{Module name} & Output size & Component \\ \hline
		\multirow{6}{*}{Stem} & Convolution & $64 \times 150 \times 150$ & $3 \times 3$ conv, stride 2 \\ \cline{2-4}
		& Convolution & $64 \times 150 \times 150$ & $1 \times 1$ conv, stride 1 \\ \cline{2-4}
		& Depth-wise convolution & $64 \times 150 \times 150$ & $3 \times 3$ dwconv, stride 1 \\ \cline{2-4}
		& Convolution & $128 \times 150 \times 150$ & $1 \times 1$ conv, stride 1 \\ \cline{2-4}
		& Depth-wise convolution & $128 \times 150 \times 150$ & $3 \times 3$ dwconv, stride 1 \\ \cline{2-4}
		& Pooling & $128 \times 75 \times 75$ & $2 \times 2$ max pool, stride 2 \\ \hline
		\multirow{10}{*}{Extractor}
		& Dense stage 0& $256 \times 75 \times 75 $  & DDB-b$(32)$ * 4 \\ \cline{2-4}
		& \multirow{2}{*}{Transition layer 0}& \multirow{2}{*}{ $ 128 \times 38 \times 38 $ } &
		$1 \times 1$ conv, stride 1 \\
		& & & $2 \times 2$ max pool, stride 2
		\\ \cline{2-4}
		& Dense stage 1 & $416 \times 38 \times 38 $  & DDB-b$(48)$ * 6 \\ \cline{2-4}
		& \multirow{2}{*}{Transition layer 1}& \multirow{2}{*}{$128 \times 19 \times 19 $ } &
		$1 \times 1$ conv, stride 1 \\
		& & & $2 \times 2$ max pool, stride 2
		\\ \cline{2-4}
		& Dense stage 2 & $512 \times 19 \times 19 $  & DDB-b$(64)$ * 6 \\ \cline{2-4}
		& Transition layer 2 & $256 \times 19 \times 19 $ & $1 \times 1$ conv, stride 1 \\ \cline{2-4}
		& Dense stage 3 & $736 \times 19 \times 19 $  & DDB-b$(80)$ * 6 \\ \cline{2-4}
		& Transition layer 3 & $64 \times 19 \times 19 $ & $1 \times 1$ conv, stride 1 \\ \hline
	\end{tabular}
	\vspace{2ex}
	\caption{Tiny-DSOD backbone architecture (input size $3\times 300\times 300$). In the "Component" column,
		the symbol "*" after block names indicates that block repeats number times given after the symbol.}\label{tab:backbone}
	\vspace{-3ex}
\end{table}

\subsection{{Depthwise FPN based Front-end}}
\begin{figure}[]
	\centering
	\footnotesize
	\includegraphics[width=0.88\textwidth]{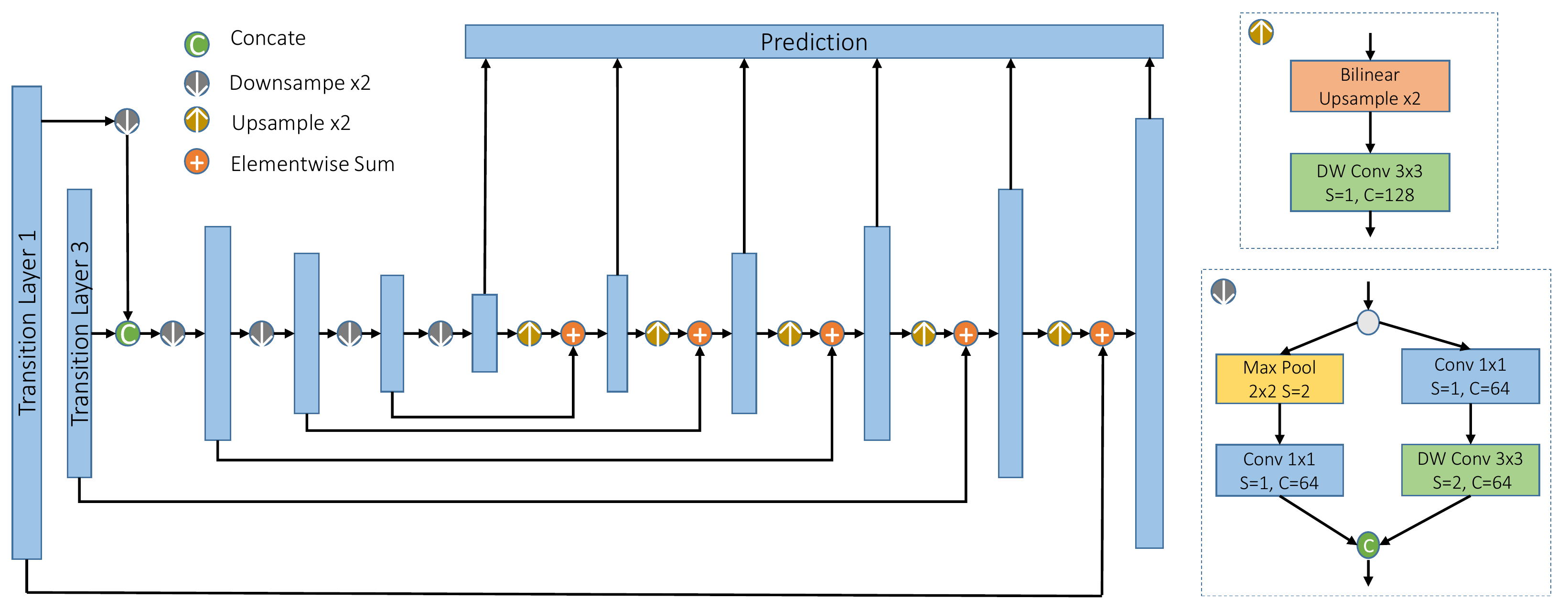}
	\vspace{1ex}
	\caption{Illustrations of the D-FPN structure.
		The left part is the over structure of D-FPN, while the right part further depicts the details of the up-sampling (top-right) and down-sampling (bottom right) modules in D-FPN. Note both sampling are by factor 2, "S" is the stride of convolution, and "C" is the number of output channels.}
	\label{fig:front}
	\vspace{-3ex}
\end{figure}

The plain structured front-end in SSD and DSOD has limitation that shallow prediction layers lack of semantic information of objects.
To overcome this problem, we borrow the feature pyramid idea from \cite{Fu2017DSSD} and \cite{Lin2017Feature}, and design a lightweight FPN named depthwise FPN (D-FPN) in our predictor to re-direct the information flow from deeper and smaller feature maps to shallower ones.
\autoref{fig:front} illustrates the structure of our front-end predictor, which consists of a downsampling path and a reverse upsampling path.
The reverse-path has been demonstrated being very helpful for small object detection in many works \cite{Xiang2017Context,Fu2017DSSD,Lin2017Feature}.
However, most of these works implement the reverse-path via deconvolution operations, which greatly increases the model complexity.
To avoid this problem, we propose a cost-efficient solution for the reverse path.
As shown in top-right of \autoref{fig:front}, we up-sample the top feature maps with a simple bilinear interpolation layer followed by a depth-wise convolution, this operation could be formulated as \autoref{eq:upsample}.
\begin{equation}\label{eq:upsample}
	F_c(x,y) = W_c * \sum_{(m,n)\in \Omega}{U_c(m,n)\tau(m, sx)\tau(n, sy)}
\end{equation}
Where $F_c$ is the $c$-th channel of output feature map and $U_c$ is the corresponding channel of input. $W_c$ is the $c$-th kernel of depth-wise convolution and $*$ denotes the spatial convolution. $\Omega$ is the co-ordinate set of input features and $s$ is the resampling coefficient in this layer. $\tau(a,b)=\max(0,1-|a-b|)$ is the differentiable bilinear operator.

The resulted feature maps are merged with the same-sized feature map in the bottom layer via element-wise addition.
Our experiment in section \ref{sec:ablation} will show that D-FPN can achieve a considerable detection accuracy boost, with slight increasing of computation cost.

\section{Experiments}\label{sec:exp}
\subsection{Implementation Details}
We implement our work on the Caffe framework~\cite{jia2014caffe}. Our model is trained from scratch with SGD solver on a server with PASCAL TitanX GPU.
Most of our training
strategies follow DSOD \cite{Shen2017DSOD}, including data augmentation, scale, L2 normalization~\cite{Liu2015ParseNet} for prediction layers, aspect ratios for default boxes, loss function (Smooth L1 loss for localization and cross entropy loss for classification) and the online hard example mining strategy.

\subsection{Ablation study on PASCAL VOC2007}\label{sec:ablation}
\noindent\textbf{\large{Design Space Exploration}}

\begin{table}[]
	\centering
	\footnotesize
	\begin{tabular}{c|c|c|c|c|c|c|c}\hline
		Row & DDB-a & DDB-b & D-FPN & Configuration & \#Params & FLOPs & mAP(\%) \\ \hline
		(1) & \checkmark & &  & G/8-8-8-24, $w$=2 & 0.90M & 1.68B & 63.1 \\
		(2) & \checkmark & &  & G/8-8-16-16, $w$=2 & 0.97M & 1.73B & 64.6 \\
		(3) & & \checkmark &  & G/32-40-56-80 & 0.82M & 0.92B & 69.3 \\
		(4) & & \checkmark &  & G/48-48-64-64 & 0.89M & 1.25B & 70.3 \\
		(5) & & \checkmark &  & G/56-56-56-56 & 0.90M & 1.35B & 70.5 \\
		(6) & & \checkmark &  & G/32-48-64-80 & 0.90M & 1.03B & 70.2 \\
		(7) & & \checkmark & \checkmark & G/32-48-64-80 & 0.95M & 1.06B & \textbf{72.1} \\ \hline
	\end{tabular}
	\vspace{2ex}
	\caption{\textit{Ablation Study on PASCAL VOC2007 test set}. The number-series G/$g_0$-$g_1$-$g_2$-$g_3$ is used to describe the network settings, where $g_i$ is the growth rate of DDB in the $i$-th stage, and $w$ is the expand ratio of DDB-a. A tick "$\checkmark$" means certain configuration is adopted in the evaluated network (row-wise), otherwise no.}\label{tab:ablation}
	\vspace{-3ex}
\end{table}

\vspace{1ex}
We first investigate the design settings in our DDB based backbone.
We conduct experiments to study two types of DDB units along with different settings of growth rate among different dense stages.
For fair comparison, we follow the common training-set settings, which we train our model on the PASCAL VOC 07+12 \verb'trainval' set, and test on the
VOC2007 \verb'test' set.
\autoref{tab:ablation} summarizes the studying results.
It shows that under similar resource usages, DDB-b based backbone performs much better than that of DDB-a.
For instance, when the parameter size is fixed to $0.90$M, DDB-b based network achieves $7.1\%$ higher mAP than that of DDB-a (70.2\% vs 63.1\%) and further saves $0.65$B FLOPs (1.03B vs 1.68B) of computing.
Hence, DDB-b is our recommended choice and used as standard setting in the following benchmark studies.

It is also obvious that the detection accuracy improves when increasing the overall growth rate in backbone. Meanwhile, we observe a trade-off between resources (parameter-size \& FLOPs) and accuracy (mAP). From row-($4$) to row-($6$) in \autoref{tab:ablation}, we find when the parameter sizes are similar, models with relative uniform growth rate will have slightly better accuracy. However, as discussed in section \ref{sec:backbone}, large growth rate in shallow stages yields high computation cost.
Therefore, we take the configuration G/32-48-64-80 (row-7) as our baseline, since it can achieve comparable accuracy with least FLOPs under the same model size constraint.

\vspace{1ex}
\noindent\textbf{\large{Effectiveness of D-FPN}}

\vspace{1ex}
We further investigate the effectiveness of our light-weighted D-FPN front-end.
Comparing last two rows in \autoref{tab:ablation}, we find D-FPN can bring $1.9\%$ performance gain with just $0.03B$ FLOPs increasing and $0.05M$ parameter-size increasing.
Such increasing in computation resources is tolerable and worthy due to noticeable accuracy gain.

\vspace{1ex}
\noindent\textbf{\large{Runtime Analysis}}

\vspace{1ex}
We compare the detection speed of Tiny-DSOD to state-of-the-art lightweight object detectors on the PASCAL VOC 2007 dataset.
The speed is measured by frame-per-second (fps) on Nvidia TitanX GPU. To accelerate the inference, we merge the parameters of batch normalization layers into the convolution operations ahead. The results are reported in the "FPS" column of \autoref{tab:voc2007}.
With 300$\times$300 input, Tiny-DSOD can process images at a speed of $9.5$ms (105 fps) with a batch size of 8, which is 4.2$\times$ faster than real-time requirement (25fps)
and is faster than other ultra-efficient detectors except Tiny-YOLO.
Our Tiny-DSOD is 6.0$\times$ faster than full-sized DSOD~\cite{Shen2017DSOD}, 2.3$\times$ faster than full-sized SSD~\cite{Liu2016SSD} and 1.5$\times$ faster than YOLOv2.
Tiny-DSOD is still slower than Tiny-YOLO, however, our model shows fewer theoretic FLOPs (see the column "FLOPs") comparing to other detectors.
The reasons are two folds.
First, Tiny-YOLO is based on plain convolution structures (without residual and concatenation), and the author made tailored GPU implementation optimization.
Second, our Tiny-DSOD directly uses Caffe without any additional optimization, where Caffe has less efficient implementation for the depthwise convolution.
We argue that when the depth-wise convolution is well implemented, our Tiny-DSOD should run at a faster speed.
Besides, we should emphasize that our Tiny-DSOD has significant fewer parameters than all the compared full-sized and lightweight detectors. Please refer to the "\#Params" column of \autoref{tab:voc2007} for more details.

\begin{table}[]
	\centering
	\footnotesize
	\begin{tabular}{c|c|c|c|c|c|c}\hline
		Method & Input size & Backbone & FPS & \#Params & FLOPs & mAP(\%) \\ \hline
		Faster-RCNN\cite{Ren2015Faster} & $600\times 1000$ & VGGNet & 7 & 134.70M & 181.12B & 73.2 \\
		R-FCN\cite{Dai2016R} & $600\times 1000$ & ResNet-50 & 11 & 31.90M & - & 77.4 \\
		SSD\cite{Liu2016SSD}$\dag$ & $300\times 300$ & VGGNet & 46 & 26.30M & 31.75B & 77.2 \\
		YOLO\cite{Redmon2016You} & $448\times 448$ & - & 45 & 188.25M & 40.19B & 63.4 \\
		YOLOv2\cite{Redmon2016YOLO9000} & $416\times 416$ & Darknet-19 & 67 & 48.20M & 34.90B & 76.8 \\
		DSOD\cite{Shen2017DSOD} & $300\times 300$ & DS/64-192-48-1 & 17.4 & 14.80M & 15.07B & 77.7 \\ \hline \hline
		Tiny-YOLO & $416\times 416$ & - & 207 & 15.12M & 6.97B & 57.1 \\
		SqueezeNet-SSD*\footnotemark[1] & $300\times 300$ & SqueezeNet & 44.7 & 5.50M & 1.18B & 64.3 \\
		MobileNet-SSD*\footnotemark[2] & $300\times 300$ & MobileNet & 59.3 & 5.50M & 1.14B & 68.0 \\
		DSOD small\cite{Shen2017DSOD} & $300\times 300$ & DS/64-64-16-1 & 27.8 & 5.90M & 5.29B & 73.6 \\
		Pelee\cite{Wang2018Pelee} & $300\times 300$ & PeleeNet & - & 5.98M & 1.21B & 70.9 \\
		Tiny-DSOD (ours) & $300\times 300$ & G/32-48-64-80 & 105 & 0.95M & 1.06B & 72.1 \\ \hline
	\end{tabular}
	\vspace{2ex}
	\caption{\textit{PASCAL VOC2007 test detection results.} "$\dag$" means the results are obtained via series of data augmentation after paper published.
		"*" means that we test those open-source models ourselves due to no speed report originally.}\label{tab:voc2007}
	\vspace{-3ex}
\end{table}
\footnotetext[1]{Models from \url{ https://github.com/chuanqi305/MobileNet-SSD}}
\footnotetext[2]{Models from \url{ https://github.com/chuanqi305/SqueezeNet-SSD}}

\subsection{Benchmark Results on PASCAL VOC2007}\label{sec:voc2007}
Our model is trained from scratch on the union of VOC2007 \verb'trainval' and VOC2012 \verb'trainval' dataset.
We use a mini batch size of 128 (accumulated over several iterations). The initial learning rate is set to 0.1, and divided by a factor of 10 every $20$k iterations.
The total number of training iterations is $100$k. We utilize a SGD solver with momentum 0.1 to optimize our objective function. Similar to \cite{Shen2017DSOD}, we use a weight decay of $0.0005$ to avoid overfitting. All our conv-layers and dwconv-layers are initialized with "xavier" method \cite{Glorot2010Understanding}.

We report the detection results on VOC2007 \verb'test' set in \autoref{tab:voc2007}, in which upper part results are from state-of-the-art full-sized detection models, while the
lower part results are from lightweight detection models.
Our Tiny-DSOD achieves 72.1\% mAP, which is significantly better than most lightweight detectors, except DSOD-smallest \cite{Shen2017DSOD}.
However, our Tiny-DSOD has only 1/6 parameters and 1/5 FLOPs to DSOD-smallest.
When comparing our model with the state-of-the-art full-sized models, there is still marginal accuracy drops.
However, Tiny-DSOD requires much smaller persistent memory for model storage and much less computing cost. For instance,
Faster-RCNN~\cite{Ren2015Faster} is just 1.1\% higher in accuracy than Tiny-DSOD, while with more than 140$\times$ larger model-size and 180$\times$ more theoretic computing cost (practically, 10$\times$ slower in fps).
These comparisons show that Tiny-DSOD achieves much better trade-off between resources (model size \& FLOPs) and detection accuracy, which is extremely useful for resource-restricted usages.

\subsection{Benchmark Results on KITTI}

Next we evaluate our detector on the autonomous driving usages for the KITTI 2D object detection task \cite{Geiger2012KITTI}.
Different from PASCAL VOC, KITTI dataset is composed of extremely wide images of size $1242\times 375$.
To avoid the vanishing of small objects, we resize the input image size to $1200\times 300$ instead of $300\times 300$.
This resolution will increase the FLOPs for our detector but will maintain good detection accuracy.
Following the configuration in \cite{Wu2017SqueezeDet}, we randomly split the 7381 images half into training set and half into validation set. The average precision is tested on the validation set. The batch size of training is set to $64$. We start our training process with a learning rate of $0.01$,
because loss oscillation is observed with large learning rate during training from scratch.
We divide the learning rate by $2$ every $10$k iterations. Our training stops at $60$k iterations as the amount of training images is small. Other settings are identical to the experiments on PASCAL VOC2007 in section \ref{sec:voc2007}.

\begin{figure}[]
	\centering
	\footnotesize
	\includegraphics[width=\textwidth]{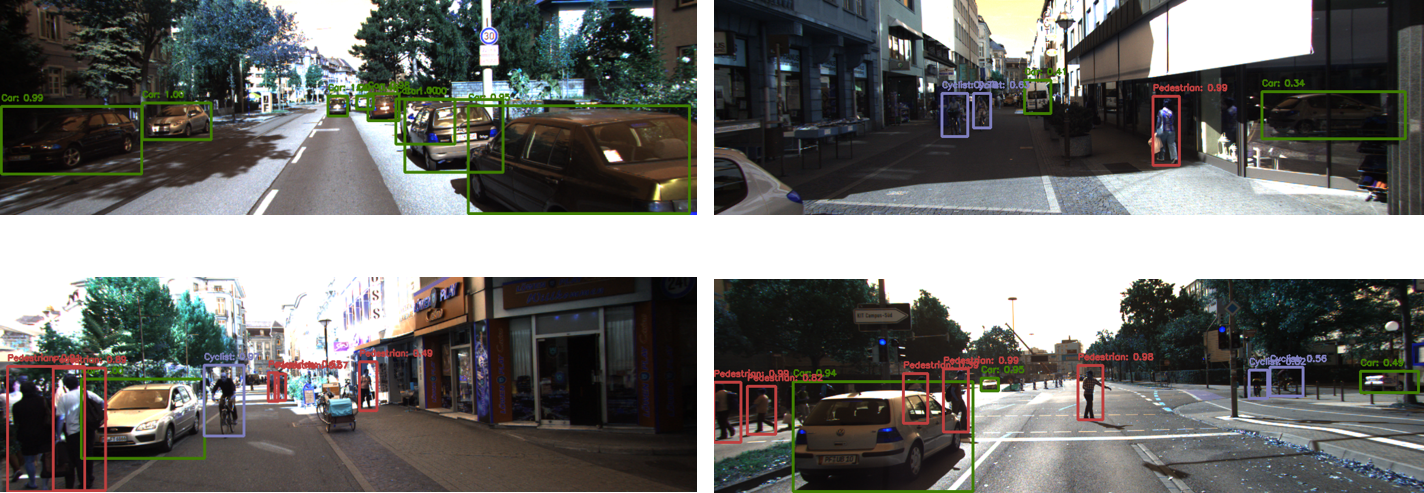}
	\vspace{3ex}
	\caption{Examples on kitti val set of road detection from the output of Tiny-DSOD. Each output bounding box is colored encoded into correponding category and filtered by a confidence threshold of 0.3 for visualization}
	\label{fig:kitti_vis}
	\vspace{-3ex}
\end{figure}

The results on validation set are reported in \autoref{tab:kitti}. Our Tiny-DSOD achieves a competitive result of 77.0\% mAP, which is slightly better than SqueezeDet \cite{Wu2017SqueezeDet} (77.0\% vs 76.7\%), while our model reduces more than $50\%$ of the model parameters and FLOPs of computing and runs at a faster run-time speed of 15ms per image (64.9 fps), indicating that Tiny-DSOD is much more efficient under this scenario. Besides, it should be noted that Tiny-DSOD achieves the highest accuracy on the "cars" category, which is the main objects in KITTI dataset.
Figure \ref{fig:kitti_vis} further illustrates some detection examples on KITTI dataset.

\begin{table}[]
	\centering
	\footnotesize
	\begin{tabular}{c|c|c|c|c|c|c|c|c} \hline
		Method & Input size & \#Params & FLOPs & Car & Cyclist & Person &  mAP(\%) & FPS \\ \hline
        SubCNN\cite{Xiang2017Subcategory} & - & - & - & 86.4 & 71.1 & 73.7 & 77.0 & 0.2 \\
		MS-CNN\cite{Cai2016A} & $1242\times 375$ & 80M & - & 85.0 & 75.2 & 75.3 & 78.5 & 2.5 \\
		FRCN\cite{Ashraf2016Shallow} & $2000\times 604$ & 121.2M & - & 86.0 & - & - & - & 1.7 \\
		ConvDet\cite{Wu2017SqueezeDet} & $1242\times 375$ & 8.78M & 61.3B & 86.7 & 80.0 & 61.5 & 76.1 & 16.6 \\
		\hline \hline
		SqueezeDet\cite{Wu2017SqueezeDet} & $1242\times 375$ & 1.98M & 9.7B & 82.9 & 76.8 & 70.4 & 76.7 & 57.2 \\
		Tiny-DSOD (ours) & $1200\times 300 $ & 0.85M & 4.1B & 88.3 & 73.6 & 69.1 & 77.0 & 64.9 \\ \hline
	\end{tabular}
	\vspace{2ex}
	\caption{\textit{KITTI 2D detection results.} Numbers under each category (car, cyclist, person) are the corresponding average precision (AP in \%). The column "mAP" is the mean AP over three categories.Note the parameter-size (0.85M) of Tiny-DSOD here is slightly different to the VOC case (0.95M) due to different number of object categories.}\label{tab:kitti}
	\vspace{-3ex}
\end{table}

\subsection{Benchmark Results on COCO}
Finally, we evaluate the performance of our method on the COCO dataset.
Following the common settings \cite{Ren2015Faster}, we train our model on \verb'trainval 35k' dataset, which is obtained by excluding $5$k images from \verb'val' set and merge the rest data into the $80$k \verb'train' set, and further evaluate our detector on the \verb'test-dev 2015' set. The batch size is set to 128. The initial learning rate is set to $0.1$ for the first 80k iterations, then divided by 10 after every $60$k iterations. The total number of training iterations is $320$k. The other training configurations are identical to the experiments on COCO in SSD~\cite{Liu2016SSD}.

The test results are summarized in \autoref{tab:coco}. Tiny-DSOD achieves $23.2\%$ mAP on the \verb'test-dev' set in the metric of AP@IOU[$0.5:0.95$], which outperforms the lightweight counterparts MobileNet-SSD (v1 \& v2) \cite{Sandler2018Inverted} and PeleeNet \cite{Wang2018Pelee}, and even outperforms the full-sized detector YOLOv2 \cite{Redmon2016YOLO9000}. Besides, Tiny-DSOD has significant small model comparing to all the listed methods in the table.
For instance, the state-of-the-art full-sized YOLOv2 has 58$\times$ larger model and 15.6$\times$ more FLOPs than Tiny-DSOD.
These comparisons verify that Tiny-DSOD is efficient yet accurate for resource-restricted object detection usages.

\begin{table}[]
	\centering
	\footnotesize
	\begin{tabular}{c|c|c|c|c|c|c}\hline
		\multirow{2}{*}{Method} & \multirow{2}{*}{Input size} & \multirow{2}{*}{FLOPs} & \multirow{2}{*}{\#Params} & \multicolumn{3}{c}{AP (\%), IOU} \\ \cline{5-7}
		& & & & $0.5$:$0.95$ & $0.5$ & $0.75$ \\ \hline
		SSD\cite{Liu2016SSD} & $300\times 300$ & 34.36B & 34.30M & 25.1 & 43.1 & 25.8 \\
		YOLOv2\cite{Redmon2016YOLO9000} & $416\times 416$ & 17.50B & 67.43M & 21.6 & 44.0 & 19.2 \\ \hline \hline
		MobileNet-SSDLite\cite{Sandler2018Inverted} & $300\times 300$ & 1.30B & 5.10M & 22.2 & - & - \\
		MobileNetv2-SSDLite\cite{Sandler2018Inverted} & $300\times 300$ & 0.80B & 4.30M & 22.1 & - & - \\
		Pelee\cite{Wang2018Pelee} & $304\times 304$ & 1.29B & 5.98M &
		22.4 & 38.3 & 22.9 \\
		Tiny-DSOD (ours) & $300\times 300$ & 1.12B & 1.15M & 23.2 & 40.4 & 22.8 \\ \hline
	\end{tabular}
	\vspace{2ex}
	\caption{\textit{COCO test-dev 2015 detection results.} Note the parameter-size (1.15M) of Tiny-DSOD here is slightly different to the VOC case (0.95M) due to different number of object categories.}\label{tab:coco}
	\vspace{-3ex}
\end{table}

\section{Conclusion}
This paper proposes the lightweight object detection method, namely Tiny-DSOD, for resource-restricted usages.
We realize a better trade-off between resources (FLOPs \& memory) and accuracy with two innovative blocks: depthwise dense blocks (DDB) and depthwise feature pyramid networks (D-FPN). We verify the effectiveness of the invented blocks and detectors through extensive ablation studies.
We compare Tiny-DSOD to state-of-the-art lightweight detectors such as MobileNet-SSD (v1 \& v2), SqueezeDet, Pelee on three object detection benchmarks (PASCAL VOC 2007, KITTI, COCO).
It shows that Tiny-DSOD outperform those methods in each benchmark on all the three metrics (accuracy, speed in term of FLOPs, and parameter-size).
Especially, Tiny-DSOD achieves 72.1\% mAP on PASCAL VOC 2007 with just 0.95M parameters and 1.14B FLOPs of computing.
This is by far the state-of-the-art result with such a low resource requirement.

\section*{Acknowledgement}
Yuxi Li and Weiyao Lin are supported by NSFC (61471235) and Shanghai "The Belt and Road" Young Scholar Exchange Grant(17510740100).

\bibliography{egbib}
\end{document}